\documentclass[10pt,twocolumn,letterpaper]{article}

\usepackage[pagenumbers]{cvpr} 
\usepackage{amssymb}

\definecolor{cvprblue}{rgb}{0.21,0.49,0.74}
\usepackage[pagebackref,breaklinks,colorlinks,allcolors=cvprblue]{hyperref}

\def\paperID{4} 
\def\confName{CVPR}
\def\confYear{2026}

\title{Form and Void: Entangled Composition through an Autonomous AI Agent}

\author{
Shiwen Wang$^{1,*}$ \quad Jian Yang$^{2,*}$ \quad Xu Wang$^{1}$ \quad Xincan Wang$^{3,\dagger}$ \quad Weiming Dong$^{1, 4,\dagger}$ \\[1.5mm]
{\small $^{1}$School of AI, University of Chinese Academy of Sciences \quad $^{2}$Renmin University of China} \\
{\small $^{3}$Shanghai Theatre Academy \quad $^{4}$MAIS, Institute of Automation, Chinese Academy of Sciences} \\[1mm]
{\tt\small wangxincan@sta.edu.cn} \quad {\tt\small weiming.dong@ia.ac.cn}
}

\begin{document}

\twocolumn[{%
\renewcommand\twocolumn[1][]{#1}%
\maketitle
\captionsetup{type=figure}
\begin{center}
\vspace{-1em}
    \includegraphics[width=\linewidth]{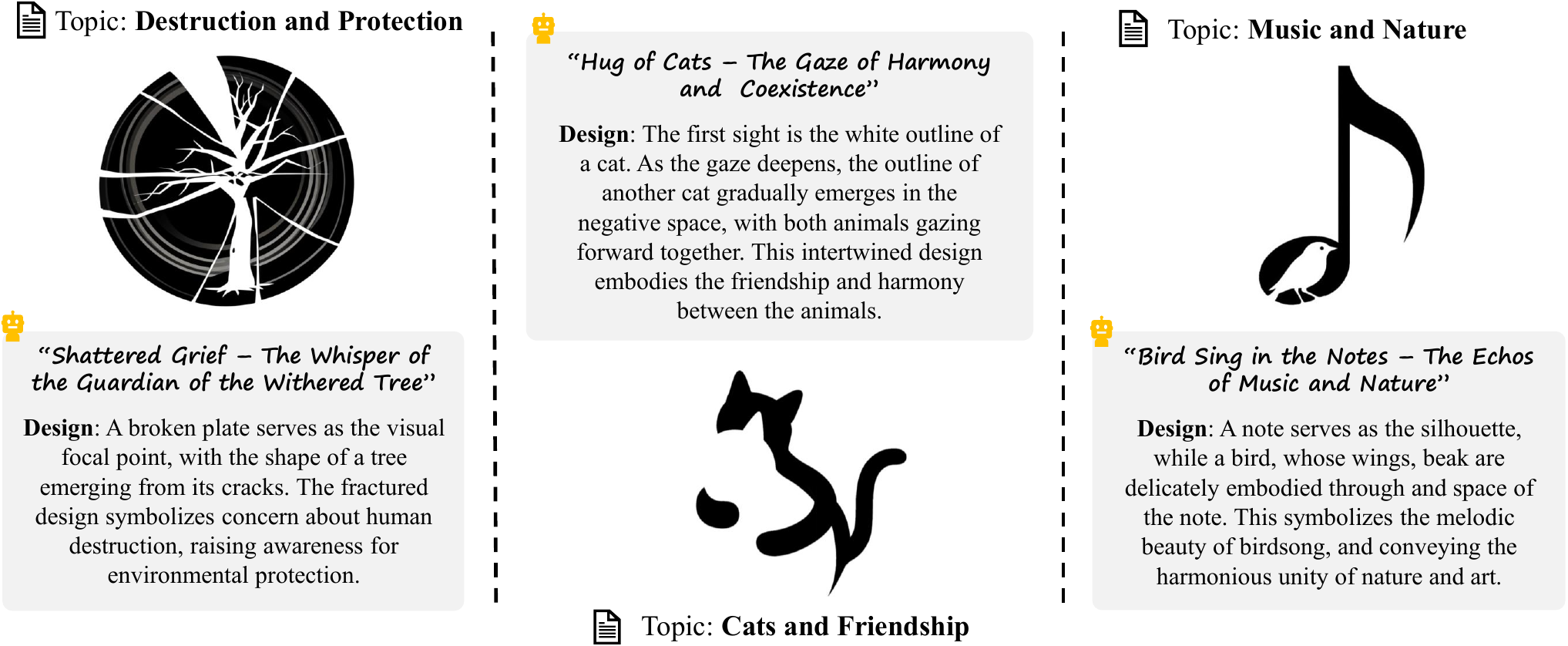}
        \caption{Given a topic, our FaV-A autonomously generates an artwork accompanied by a caption and a detailed description, both grounded in the entangled duality ``Form and Void''. The first two lines within the shadow box present the artwork's caption. The ``Design'' section offers a description of the composition and the layered meanings embedded within the piece.}
    \label{fig:teaser}
    \end{center}%
}]

\begingroup
\renewcommand{\thefootnote}{}
\makeatletter
\let\Hy@raisedlink\@gobble
\makeatother
\footnotetext{$^*$Equal contribution. $^\dagger$Corresponding authors.}
\endgroup

\begin{abstract}
Positive and negative space is a fundamental principle in visual composition, supporting visually coherent forms and layered semantic relationships. Generating such compositions is challenging because it requires coordinated control over two semantic concepts that share a common boundary. Although recent text-to-image models and multimodal large language models (MLLMs) have achieved strong performance in image generation and visual understanding, positive-negative space generation remains difficult, particularly under direct single-pass prompting. In this work, we present the \textbf{F}orm \textbf{a}nd \textbf{V}oid \textbf{A}gent (\textbf{FaV-A}), a multimodal agent designed for staged positive-negative space generation. FaV-A follows a progressive workflow: it first generates a base object, then analyzes its shape and spatial structure to identify candidate negative-space semantics, and finally produces compositional instructions for the final image generation stage. Experimental results and ablation analyses suggest that FaV-A provides a more effective framework than direct zero-shot MLLM baselines for producing visually coherent and semantically aligned positive-negative space compositions.
\end{abstract}

\section{Introduction}

The interplay between form and void lies at the core of visual composition and is classically expressed through positive and negative space. In art and design, negative space refers to the empty regions around and between the subjects of an image~\cite{lupton2015graphic, lucie-smith1984negative, cave2013shooting}. This principle plays an important role across visual art, photography~\cite{photography}, graphic design~\cite{graphicdesign}, and sculpture~\cite{sculpture}. Designing with negative space requires careful coordination of visual elements, as figure-ground relationships affect not only compositional structure but also semantic interpretation. From a computational perspective, generating such compositions is challenging because the model must simultaneously represent an explicit foreground form and an implicit secondary shape induced by the surrounding space.

Recent text-to-image models have shown strong capabilities in visual generation. However, existing generation pipelines often rely on explicit spatial controls, such as masks, layouts, or other manually specified conditions, to manage figure-ground relationships. With the emergence of multimodal large language models (MLLMs)~\cite{openai2023, bagel,gemini}, visual understanding has improved substantially. Nevertheless, a gap remains between understanding and generation: although MLLMs can often recognize and describe positive-negative space compositions, generating such effects in a zero-shot, single-pass setting remains difficult. In our preliminary observations, directly prompting an end-to-end model to synthesize two overlapping semantic concepts with a shared boundary often results in structurally unstable compositions or semantically ambiguous outputs. These observations motivate a staged generation framework in which structure discovery and semantic refinement are separated into successive steps. 

To address this gap, we adopt a staged generation perspective motivated by common design workflows. In positive-negative space composition, a plausible workflow is to first establish a primary subject, then inspect its contour and spatial structure, and finally refine the image by introducing secondary semantics in the negative space. Following this intuition, we model the task as a sequence of intermediate steps in which earlier outputs provide structural guidance for later stages. This decomposition reduces the burden of requiring a model to resolve shared-boundary geometry and dual semantic alignment in a single generation step. In addition, recent work on MLLM-based agents suggests that explicit intermediate reasoning can be beneficial for complex multimodal generation tasks~\cite{feng2023layoutgpt,gao2025gllava,fang2025got}.

Motivated by this perspective, we present the Form and Void Agent (FaV-A), a multimodal agent for positive-negative space generation from abstract topics. Rather than relying on one-shot synthesis, FaV-A adopts a staged workflow. Given a topic, it first generates a base object informed by design-oriented priors. It then analyzes the base image to identify shape and spatial cues that could support a secondary interpretation in the negative space. Finally, these observations are translated into compositional instructions for the final image generation step. As shown in Figure~\ref{fig:teaser}, this staged process yields more coherent positive-negative space compositions in our examples. These results suggest that multimodal agents are a promising direction for design-oriented generation tasks that require coordinated spatial and semantic control.

To summarize, this work makes three contributions. First, we formulate positive-negative space generation as a challenging multimodal design problem that requires coordinated spatial and semantic reasoning under shared-boundary constraints. Second, we propose the Form and Void Agent (FaV-A), a staged multimodal agent that decomposes this task into three successive phases: base-object generation, image-aware structural analysis, and instruction-guided final synthesis. Third, through qualitative comparisons, user evaluation, and ablation analysis, we show that this staged framework produces more structurally coherent and semantically aligned positive-negative space compositions than direct zero-shot MLLM prompting.

\section{Related Work}
\label{sec:related_work}

\subsection{Text to Image Generation}
Recent diffusion-based methods have shown that generative models can produce images with multiple valid interpretations under predefined transformations or viewing setups, such as multi-view optical illusions or physically arranged illusion patterns~\cite{geng2024visualanagrams,burgert2023diffusionillusions}. However, these methods typically target fixed transformation rules or illusion mechanisms, rather than discovering a figure-ground composition from an abstract semantic topic.

Controllable image generation with explicit structural conditions is also the research focus. Methods such as ControlNet, T2I-Adapter, and GLIGEN enable text-to-image models to follow masks, edges, depth maps, keypoints, bounding boxes, or other grounding signals~\cite{zhang2023controlnet,mou2024t2iadapter,li2023gligen}. These approaches provide strong spatial control once the control signal is available, but they generally assume that the relevant structure has already been specified. In contrast, our setting begins with an abstract topic and requires the system to infer both a suitable primary form and a compatible negative-space interpretation without manually designed spatial templates.

\subsection{MLLM Agents for Visual Generation}
Recent MLLM-based agents have demonstrated strong capabilities in visually grounded planning and execution~\cite{koh2024visualwebarena}. In creative and design-oriented settings, PosterLLaVA formulates poster layout generation with multimodal large language models, GenArtist uses an MLLM agent to coordinate image generation and editing through planning and verification, and BannerAgency applies multimodal agents to the creation of editable advertising banners~\cite{yang2024posterllava,wang2024genartist,wang2025banneragency}. GraphicBench further shows that graphic design remains challenging for language agents, especially when reasoning about spatial relationships and coordinating global dependencies across design decisions~\cite{ki2025graphicbench}.

These works motivate the use of staged multimodal reasoning for design tasks, but they primarily focus on layout planning, tool orchestration, or editable asset generation. Our work instead targets positive-negative space synthesis, where the central challenge is to derive a shared-boundary figure-ground composition from an abstract topic through intermediate image analysis and compositional anchoring.
\section{Method}
\begin{figure*}[!htp]
\vspace{-0.5cm}
    \centering
    \includegraphics[width=\linewidth]{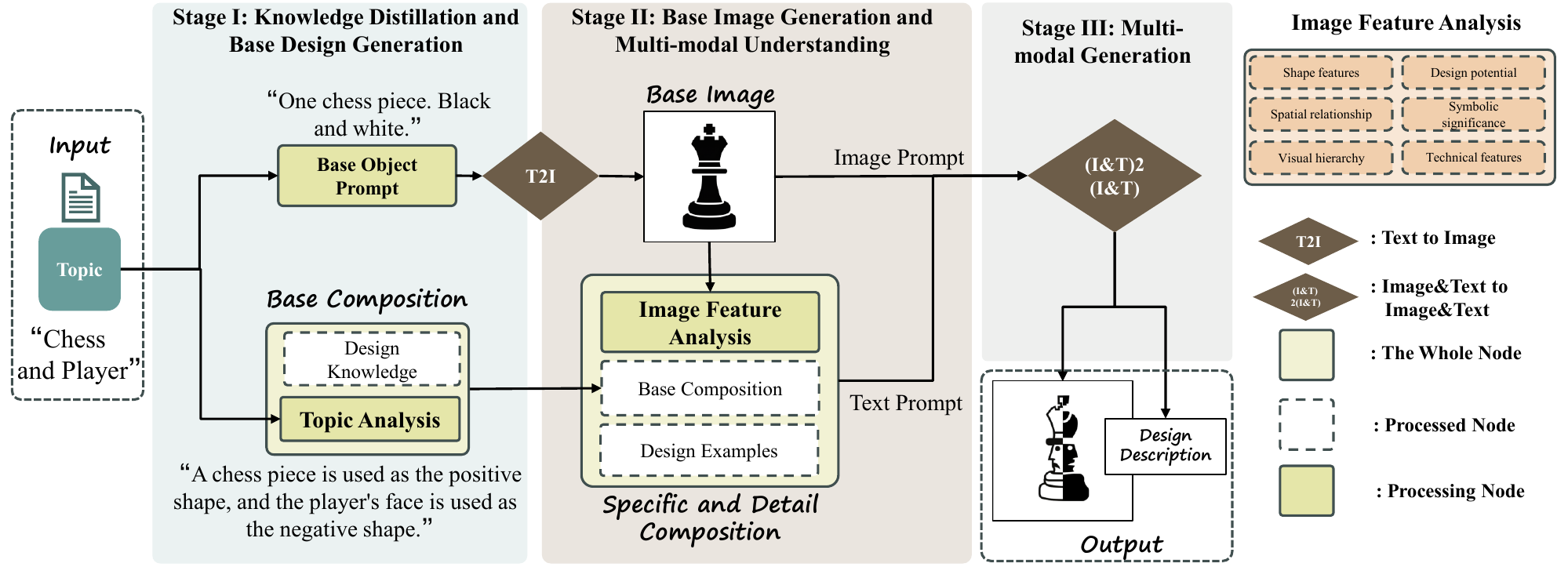}
    \caption{Multistage design process of FaV-A to create positive-negative space compositions.}
    \label{fig:method}
    \vspace{-0.5cm}
\end{figure*}
\label{sec:method}

\subsection{Overall Architecture}
\label{sec:method_formulation}

We propose the Form and Void Agent (FaV-A), a multimodal agent for positive-negative space generation built on a large multimodal model, denoted by $\mathcal{M}$. Given an abstract textual topic $\mathcal{T}$, the goal of FaV-A is to generate a final composition $\mathbf{I}_{\mathrm{final}}$ together with a textual description $\mathcal{D}_{\mathrm{final}}$ that captures the intended figure-ground duality. 

As illustrated in Figure~\ref{fig:method}, unlike layout-controlled generation pipelines that depend on manually specified masks, edges, or spatial layouts, FaV-A follows a staged and mask-free generation process. Instead of directly mapping $\mathcal{T}$ to the final image in a single step, the agent decomposes the task into three sequential stages. In Stage I, Topic-Driven Semantic Planning ($\mathcal{M}_{\mathrm{plan}}$), the agent converts the input topic into a base-object prompt and a high-level compositional plan, guided by design priors $\mathcal{P}_{\mathrm{prior}}$. In Stage II, Multi-Modal Spatial Parsing and In-Context Anchoring ($\mathcal{M}_{\mathrm{parse}}$), the agent generates a base image and analyzes its shape and spatial structure to derive constraints for negative-space composition. In Stage III, Mask-Free Entangled Synthesis ($\mathcal{M}_{\mathrm{syn}}$), the agent uses the base image and the derived structural cues to produce the final positive-negative space composition.

This staged generation process can be written as:
\begin{equation}
(\mathbf{I}_{\mathrm{final}}, \mathcal{D}_{\mathrm{final}})
=
\mathcal{M}_{\mathrm{syn}}
\Big(
\mathbf{I}_{\mathrm{base}},
\mathbf{A}
\Big),
\end{equation}
\begin{equation}
(\mathbf{I}_{\mathrm{base}}, \mathbf{A})
=
\mathcal{M}_{\mathrm{parse}}
\Big(
\mathcal{M}_{\mathrm{plan}}(\mathcal{T}, \mathcal{P}_{\mathrm{prior}})
\Big),
\end{equation}
where $\mathbf{A}$ denotes the spatial and compositional cues extracted from the intermediate analysis stage.

\subsection{Stage I: Semantic Planning}
\label{sec:method_stage1}

Abstract topics $\mathcal{T}$ usually lack explicit geometric and spatial specifications. The goal of Stage I is therefore to transform the input topic into structured textual guidance for later generation stages. We provide the planning model with a system prompt containing design priors $\mathcal{P}_{\mathrm{prior}}$, which encourage attention to figure-ground structure, negative space, and compositional balance. The prompts, design priors and detailed pipeline will be included in our Supplementary Material.

Conditioned on $\mathcal{T}$ and $\mathcal{P}_{\mathrm{prior}}$, the planning model outputs:
\begin{equation}
    (\mathbf{t}_{\mathrm{obj}}, \mathbf{c}_{\mathrm{base}}) = \mathcal{M}_{\mathrm{plan}}(\mathcal{T}, \mathcal{P}_{\mathrm{prior}}),
\label{eq:stage1}
\end{equation}
where $\mathbf{t}_{\mathrm{obj}}$ denotes the \textit{Base Object Prompt} for the primary subject, and $\mathbf{c}_{\mathrm{base}}$ denotes the \textit{Base Composition Blueprint} describing the intended figure-ground arrangement. This stage converts an abstract topic into a more concrete semantic specification for subsequent image generation and analysis.

\subsection{Stage II: Spatial Anchoring}
\label{sec:method_stage2}

Given the intermediate plan from Stage I, a text-to-image module first generates a base image,
\begin{equation}
    \mathbf{I}_{\mathrm{base}} = \mathcal{G}_{\mathrm{T2I}}(\mathbf{t}_{\mathrm{obj}}).
\end{equation}
This base image provides an initial foreground structure, but it does not yet encode the intended positive-negative space composition. To derive guidance for the final synthesis stage, FaV-A performs a multimodal analysis of $\mathbf{I}_{\mathrm{base}}$.

Specifically, the agent analyzes the base image along several dimensions, including shape attributes, spatial relationships, visual hierarchy, potential negative-space interpretations, symbolic associations, and implementation-related considerations. We denote the resulting intermediate analysis by
\begin{equation}
    \mathbf{S} = \Phi(\mathbf{I}_{\mathrm{base}}),
\end{equation}
where $\mathbf{S}$ summarizes the structural and semantic observations extracted from the base image.

To further align this analysis with the target composition, FaV-A incorporates multimodal in-context learning (ICL). The agent is provided with $N=3$ visual-textual exemplars,
\[
\mathcal{E} = \{(\mathbf{I}^{(i)}_{\mathrm{exam}}, \mathcal{D}^{(i)}_{\mathrm{exam}})\}_{i=1}^N,
\]
together with the base composition blueprint $\mathbf{c}_{\mathrm{base}}$ and the parsed representation $\mathbf{S}$. Based on these inputs, the parsing module generates a detailed compositional prompt:
\begin{equation}
    \mathcal{P}_{\mathrm{anchor}} =
    \mathcal{M}_{\mathrm{parse}}(\mathbf{I}_{\mathrm{base}}, \mathbf{S}, \mathbf{c}_{\mathrm{base}}, \mathcal{E}),
\label{eq:stage2}
\end{equation}
where the resulting prompt $\mathcal{P}_{\mathrm{anchor}}$ serves as a semantic and structural anchor for the final generation stage. Rather than specifying explicit spatial masks, it describes the intended contour relationships, negative-space interpretation, and compositional constraints in natural language.

\subsection{Stage III: Guided Final Synthesis}
\label{sec:method_stage3}

The final stage generates the target positive-negative space composition from the intermediate outputs of the previous stages. Rather than relying on manually specified bounding boxes, masks, or edge-based control signals, FaV-A conditions the synthesis process on both the base image and a detailed compositional prompt.

Formally, the synthesis stage is defined as
\begin{equation}
    (\mathbf{I}_{\mathrm{final}}, \mathcal{D}_{\mathrm{final}}) = \mathcal{M}_{\mathrm{syn}}(\mathbf{I}_{\mathrm{base}}, \mathcal{P}_{\mathrm{anchor}}),
\label{eq:stage3}
\end{equation}
where $\mathbf{I}_{\mathrm{base}}$ provides the initial visual structure and $\mathcal{P}_{\mathrm{anchor}}$ provides text-based guidance for the intended figure-ground composition. The output image $\mathbf{I}_{\mathrm{final}}$ is the final synthesized result, while $\mathcal{D}_{\mathrm{final}}$ is a textual description of the semantic relationship expressed in the composition.

This stage integrates the intermediate structural analysis and the compositional prompt into the final generation process, enabling mask-free positive-negative space synthesis under multimodal conditioning.
\section{Results and Discussions}
\label{sec:results}

\subsection{Comparison with State-of-the-Art Baselines}
\label{sec:baseline_comparison}

To evaluate the effectiveness of our multi-stage architecture, we compare FaV-A against direct prompting baselines utilizing state-of-the-art closed-source multimodal models, specifically Gemini (Gemini 3.1~\cite{gemini2026}), GPT (gpt-image-1.5~\cite{openai2025gptimage}), and Claude (Sonnet 4.6~\cite{anthropic2026claude}). We conducted all experiments using \texttt{gemini-3.1-flash}~\cite{gemini2026flashlite} as the understanding (MLLM) model and \texttt{gemini-3.1-flash-image}~\cite{gemini2026flashimage} as the image generation model. In this setup, the baseline models are provided with the target abstract topic alongside explicit theoretical definitions and principles of positive-negative space design. Concretely, all closed-source baselines are prompted with the same instruction:
\begin{quote}
\small
\textit{Definition of Positive-Negative Space Art (Form and Void): Positive-negative space refers to a composition where the positive shape (the subject) and the negative space (the area around it) are equally important and interdependent. In advanced design, they are ``entangled''---meaning they share a common, single contour line. The boundary of the positive form must perfectly define the shape of another meaningful element in the negative space, creating a visual pun or dual meaning (figure-ground reversal).}

\textit{Based on the definition above, please autonomously design and generate a minimalist black-and-white artwork for the topic: ``$\{Topic\}$''.}
\end{quote}
In this way, the baselines are given access to the same high-level conceptual knowledge as FaV-A during its planning stage. 

\vspace{-1em}
\begin{figure*}
    \centering
    \includegraphics[width=\linewidth]{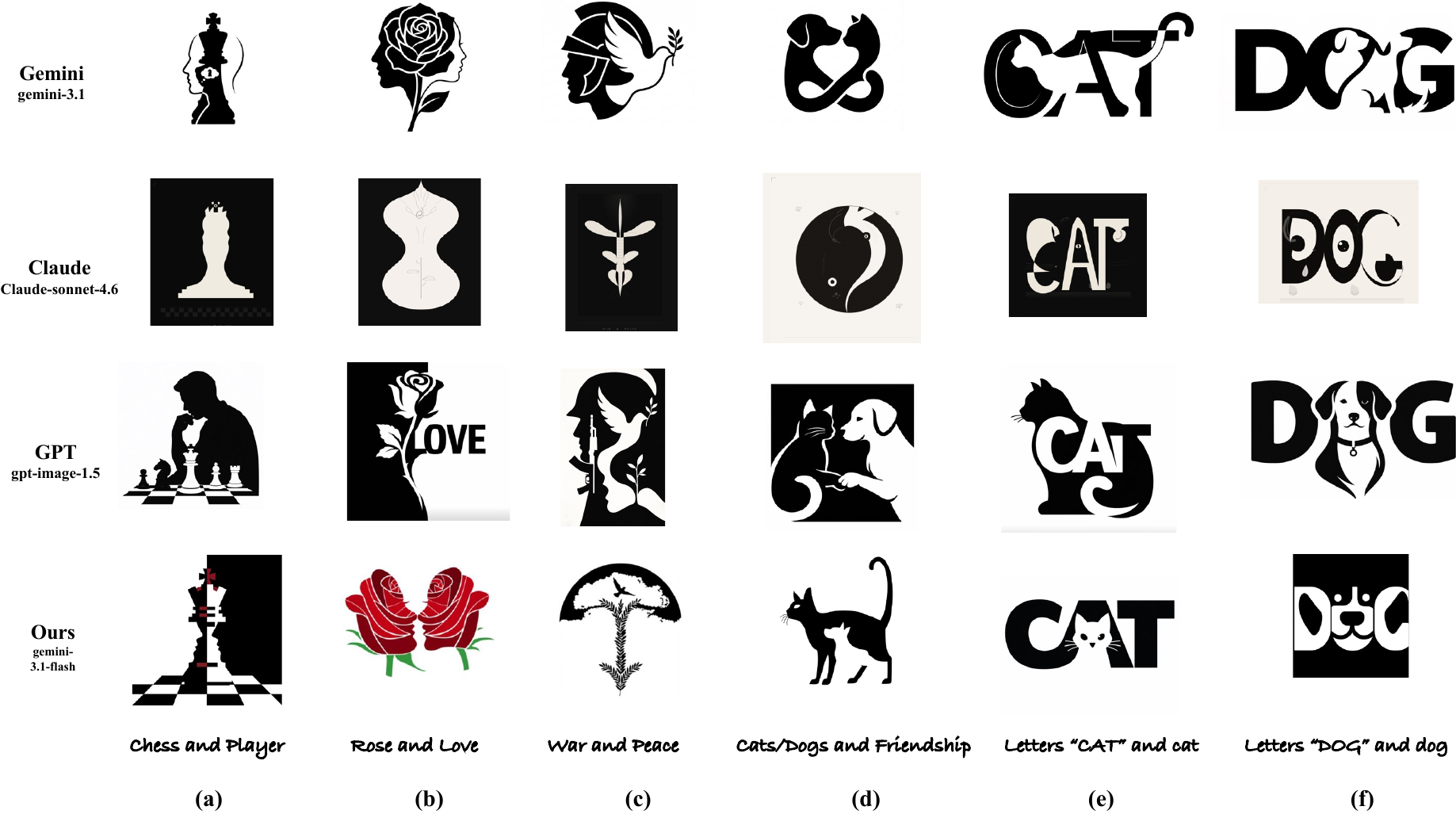} 
    \caption{Visual comparison between state-of-the-art baselines (Gemini, Claude, and GPT) and our FaV-A. Given the same topic and positive-negative space design knowledge, the baselines fail to achieve true topological entanglement: Claude defaults to semantic juxtaposition, GPT relies on explicit spatial partitioning, 
    and Gemini resorts to internal part-filling. Only FaV-A successfully entangles both subjects along a continuous shared boundary, achieving true bistable perception.}
    \label{fig:baseline_comparison}
\end{figure*}
\vspace{-1em}
\paragraph{Behavior and Limitations of Baseline Models.} 
When provided with the augmented prompt (topic + design knowledge), the baseline models are generally able to capture the target concepts and produce high-contrast black-and-white images. Nevertheless, a clear gap remains between semantic understanding and positive-negative space execution. As shown in Figure~\ref{fig:baseline_comparison}, rather than forming two mutually defined shapes through a shared boundary, the baselines exhibit three representative failure patterns: 
\begin{itemize}
    \item \textbf{Semantic Drift and Weak Topic Grounding (Claude):}  
    Claude exhibits the weakest alignment with the intended topics. Although it occasionally produces clean black-and-white graphics, its outputs often fail to preserve the core semantic concepts specified in the prompt. This issue is particularly evident in \textit{Rose and Love} and \textit{War and Peace}, where few clearly recognizable topic-related elements remain. Hence, Claude's limitation is not only its inability to realize positive-negative space structure, but also its insufficient topic grounding. As a result, its outputs lack both semantic fidelity and the dual-form coupling required for effective positive-negative space design.

    \item \textbf{Topic-Consistent but Additive Composition (Gemini):}  
    Gemini generally preserves the target concepts and produces recognizable topic-relevant objects. However, its dominant strategy is additive composition, typically by stacking, juxtaposing, or locally inserting one object into another. While semantically consistent, such layouts rarely achieve the figure-ground coupling central to positive-negative space art. Instead, the generated forms remain illustrative rather than structurally interdependent. Consequently, Gemini produces coherent topic-driven graphics, but lacks the contour sharing, ambiguity, and formal tension characteristic of strong positive-negative space compositions.

    \item \textbf{Graphic Plausibility without True Figure-Ground Reversal (GPT):}  
    GPT shows the strongest overall design sensibility among the direct-prompting baselines. Relative to simple object juxtaposition, it more often yields compositions with clearer visual hierarchy, stronger symbolic organization, and better poster-like readability. However, these results are better characterized as literal graphic design than genuine positive-negative space design. Multiple concepts are typically integrated through explicit depiction or layout-level composition, rather than being jointly defined by a shared contour across positive and negative regions. Therefore, although GPT approaches visually compelling conceptual design, it still falls short of the stricter figure-ground entanglement required in positive-negative space art.
\end{itemize}

\noindent\textbf{Advantages of the FaV-A Architecture.} 
In contrast, our proposed method successfully achieves structural entanglement by reformulating the visual illusion as a progressive, constrained reasoning task. Rather than expecting an MLLM to perform zero-shot spatial alignment in a single generation step, FaV-A utilizes a base image to establish an initial geometric foundation. By conducting multi-dimensional visual parsing on this base image, the agent synthesizes a precise semantic spatial anchor ($\mathcal{P}_{\mathrm{anchor}}$). This anchor explicitly maps the semantic requirements to specific topological regions (e.g., instructing the model to align a facial profile with the jagged external edge of a primary object). Guided by these rigorous spatial constraints, the final synthesis stage aligns the conflicting visual priors, forcing the positive and negative spaces to share a single, continuous contour. This methodology ensures the reliable emergence of bistable perception, realizing the ``one shape, dual meaning'' principle that direct prompting baselines cannot achieve.

\begin{figure*}[t]
    \centering
    \vspace{-0.7cm}
    \includegraphics[width=0.96\linewidth]{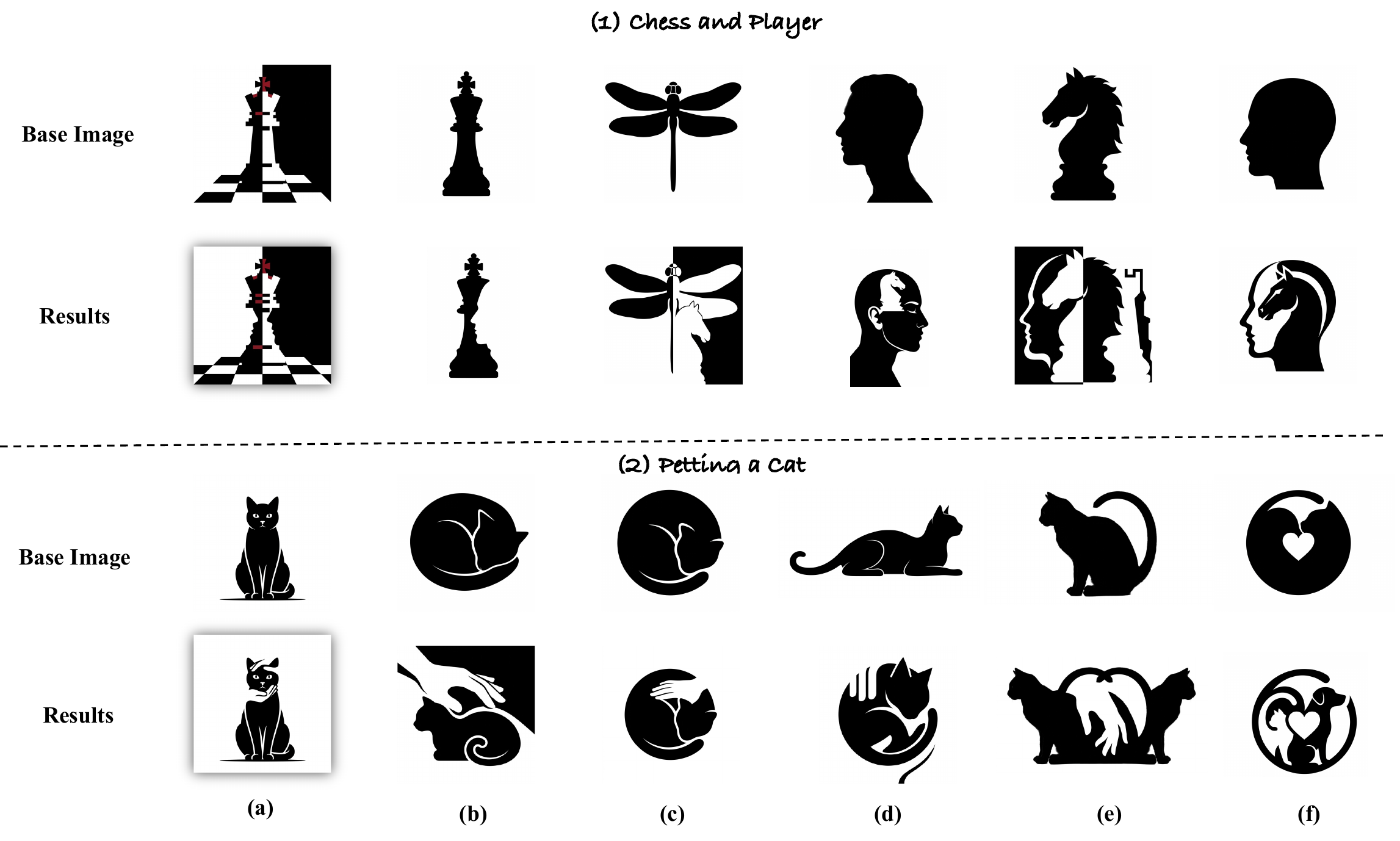}
    \caption{Ablation study on the ``Chess and Player'' (top) and ``Petting a Cat'' (bottom) topics. (a) \textbf{Full FaV-A Pipeline}, showcasing strong figure-ground interplay and shared boundaries; (b) \textbf{w/o Topic Analysis}, resulting in a shallow vertical juxtaposition; (c) \textbf{w/o Detailed Base Prompt}, introducing early semantic deviation (e.g., generating a dragonfly instead of a player); (d) \textbf{w/o Image Feature Analysis}, leading to a loss of structural fidelity and morphological drift; (e) \textbf{w/o Specific Composition}, causing visual confusion and semantic bleeding; (f) \textbf{w/o Multi-modal Generation}, leading to complete structural dissolution (e.g., the cat arbitrarily transforms into a dog).}
    \label{fig:ablation}
    \vspace{-0.5cm}
\end{figure*}

\begin{table}[h]
    \centering
    \caption{Quantitative results of the ablation study based on human evaluation ($N=50$). Metrics are evaluated on a 5-point Likert scale (higher is better). Our full FaV-A pipeline significantly outperforms all ablated variants across Topic Adherence (TA), Gestalt Quality (GQ), and overall Aesthetic Score (AS).}
    \label{tab:ablation_quant}
    \resizebox{\linewidth}{!}{
    \begin{tabular}{lccc}
    \hline
    Model Variant & TA $\uparrow$ & GQ $\uparrow$ & AS $\uparrow$ \\
    \hline
    w/o Topic Analysis & $1.35 \pm 0.42$ & $3.12 \pm 0.65$ & $2.85 \pm 0.70$ \\
    w/o Detailed Base Prompt & $4.15 \pm 0.50$ & $2.90 \pm 0.58$ & $3.25 \pm 0.62$ \\
    w/o Image Feature Analysis & $3.80 \pm 0.55$ & $2.15 \pm 0.60$ & $2.40 \pm 0.65$ \\
    w/o Specific Composition & $3.92 \pm 0.48$ & $1.85 \pm 0.52$ & $2.10 \pm 0.55$ \\
    w/o Multi-modal Generation & $3.45 \pm 0.60$ & $1.60 \pm 0.45$ & $1.80 \pm 0.48$ \\
    \hline
    \textbf{FaV-A (Ours - Full)} & \textbf{4.65 $\pm$ 0.35} & \textbf{4.58 $\pm$ 0.40} & \textbf{4.52 $\pm$ 0.38} \\
    \hline
    \end{tabular}
    }
    \vspace{-1em}
\end{table}

\paragraph{User Study} 
To rigorously quantify these qualitative observations, we conducted a user study ($N=50$) to evaluate the ablated variants across three dimensions: Topic Adherence (TA), Gestalt Quality (GQ), and overall Aesthetic Score (AS) on a 5-point Likert scale. The participants were recruited from graduate students and researchers with prior experience in visual design, computer vision, or digital art, including a subset with formal design training. During evaluation, all results were anonymized and presented in randomized order, such that the raters were unaware of which method produced each image. Each participant was asked to score the generated artworks independently based on the three criteria, following a unified instruction sheet describing the meaning of each metric. This procedure was adopted to reduce presentation bias and improve the reliability of subjective assessment.

As reported in Table~\ref{tab:ablation_quant}, the full FaV-A pipeline significantly outperforms all ablated variants. The quantitative results strongly align with our visual findings: notably, the removal of topic analysis causes a severe drop in TA (1.35), reflecting the model's inability to follow the semantic prompt. Meanwhile, the absence of specific compositional constraints (GQ: 1.85) and the bypass of multi-modal generation (GQ: 1.60) incur the heaviest penalties in Gestalt Quality, statistically confirming their indispensability in maintaining the structural illusion.

\subsection{Ablation Study}
\label{sec:ablation}

To validate the effectiveness of each functional module within the FaV-A pipeline, we conduct a set of ablation studies using the abstract topics ``Chess and Player'' and ``Petting a Cat''. For each configuration, we remove or simplify a single component while keeping the remainder of the pipeline unchanged. Figure~\ref{fig:ablation} visualizes the qualitative degradation in structural coherence and positive-negative space entanglement when individual stages are ablated.

\vspace{1mm}\noindent\textbf{Topic Analysis.} 
The primary role of this module is to decompose an abstract concept into candidate positive and negative entities and establish their initial spatial relationship. For the ablation, we bypass the semantic planning and feed the raw topic directly to the subsequent stages. The absence of this preliminary logic causes the output to degenerate into a shallow juxtaposition of elements (see Figure~\ref{fig:ablation}(b)). In the ``Petting a Cat'' example, the design merely stacks a hand silhouette and a cat silhouette vertically, lacking intentional interaction. In contrast, the full pipeline allows the hand to act as a negative shape that wraps around and gently strokes the cat's head, demonstrating a much stronger figure-ground interplay (see Figure~\ref{fig:ablation}(a)).

\vspace{1mm}\noindent\textbf{Base Object Prompt.} 
This module translates the topic into morphological instructions to synthesize a base image capable of supporting negative space carving. For the ablation, we replace the LLM-derived instructions with a generic manual prompt: \textit{``Positive or negative shapes representing \{topic\}''}. As observed in the upper example of Figure~\ref{fig:ablation}(c), this underspecified prompt introduces severe semantic deviation at the source---generating an irrelevant dragonfly-like shape due to an unstable interpretation of the word ``player''. This error propagates through the pipeline, proving that precise initial morphological constraints are essential to prevent downstream semantic drift.

\vspace{1mm}\noindent\textbf{Image Feature Analysis.} 
This stage grounds the spatial reasoning in the actual geometry (e.g., shape characteristics, visual hierarchy) of the generated base image. As illustrated in Figure~\ref{fig:ablation}(d), removing this multi-dimensional parsing disconnects the refinement process from the original base object. Consequently, the final outputs suffer from a loss of structural fidelity: in the upper example, the player's hair characteristics are noticeably altered; in the lower example, the cat's pose drifts. This analysis is particularly crucial for a human-in-the-loop workflow, where the intermediate base form selected by a designer must remain morphologically stable during final generation.

\vspace{1mm}\noindent\textbf{Specific Composition.} 
The purpose of this component is to synthesize a localized spatial anchor ($\mathcal{P}_{\mathrm{anchor}}$) that dictates exactly how the contours should be sculpted. We ablate this by substituting the detailed anchor with a naive instruction: \textit{``Transform this into a POSITIVE-NEGATIVE SPACE poster. CRITICAL: Use ONLY pure black and pure white...''} As shown in Figure~\ref{fig:ablation}(e), without the detailed topological constraints, the model exhibits morphological drift and weakened visual harmony. The positive and negative shapes become visually confused (e.g., overlapping messily in the chess example) or fail to preserve the intended object configuration (e.g., the number of cats inexplicably changes in the lower example).

\vspace{1mm}\noindent\textbf{Multi-modal Generation.} 
The final \texttt{(I\&T) to (I\&T)} stage enforces high-frequency edge preservation by integrating the accumulated visual and textual priors. By bypassing this dedicated reasoning pass and directly feeding the intermediate inputs to a standard image generator (Figure~\ref{fig:ablation}f), the system fails to consolidate the spatial constraints. This results in structural dissolution, where the visual illusion collapses entirely. Most notably, in the lower example, the cat is arbitrarily transformed into a dog, indicating that without this final reasoning mechanism, the shape priors established in earlier stages are completely ignored.

\begin{figure*}[t]
\vspace{-1cm}
\subfloat[]{
    \centering
    \includegraphics[width=0.93\linewidth]{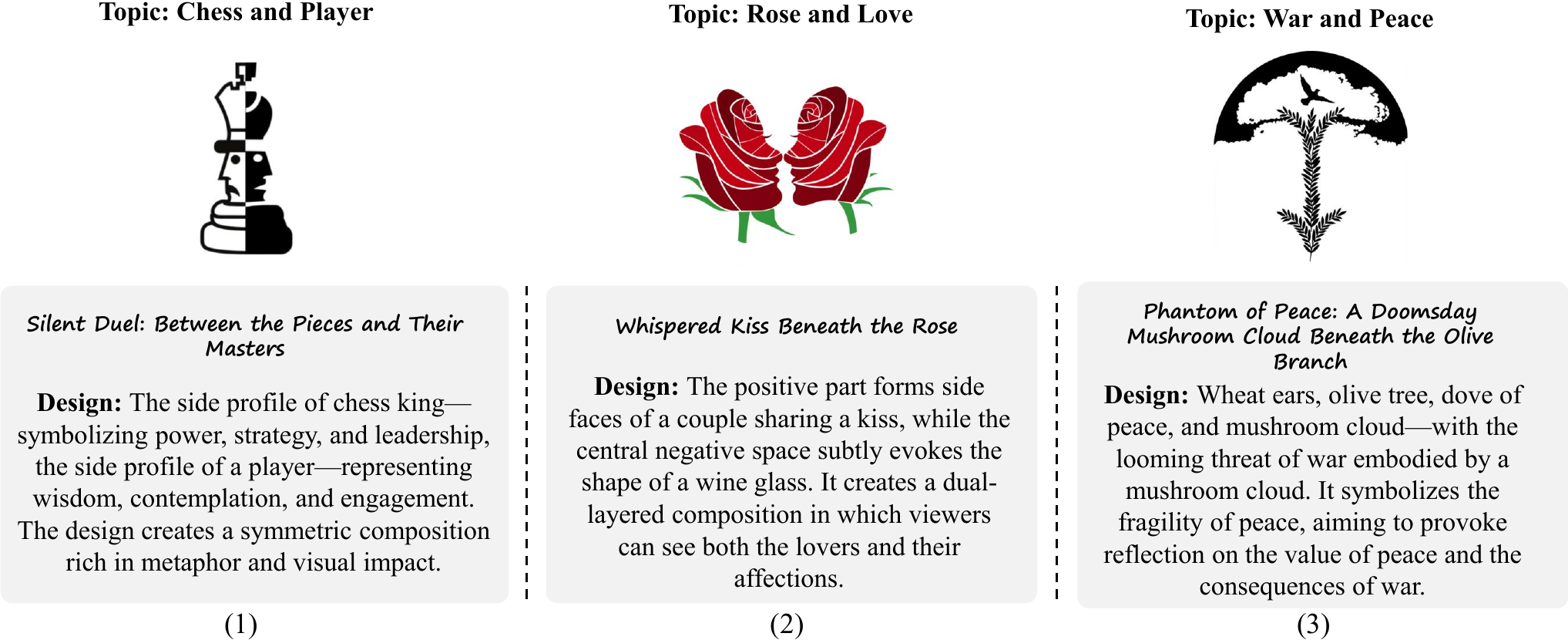}
    \label{fig:r1}
}\\
\subfloat[]{
    \centering
    \includegraphics[width=0.93\linewidth]{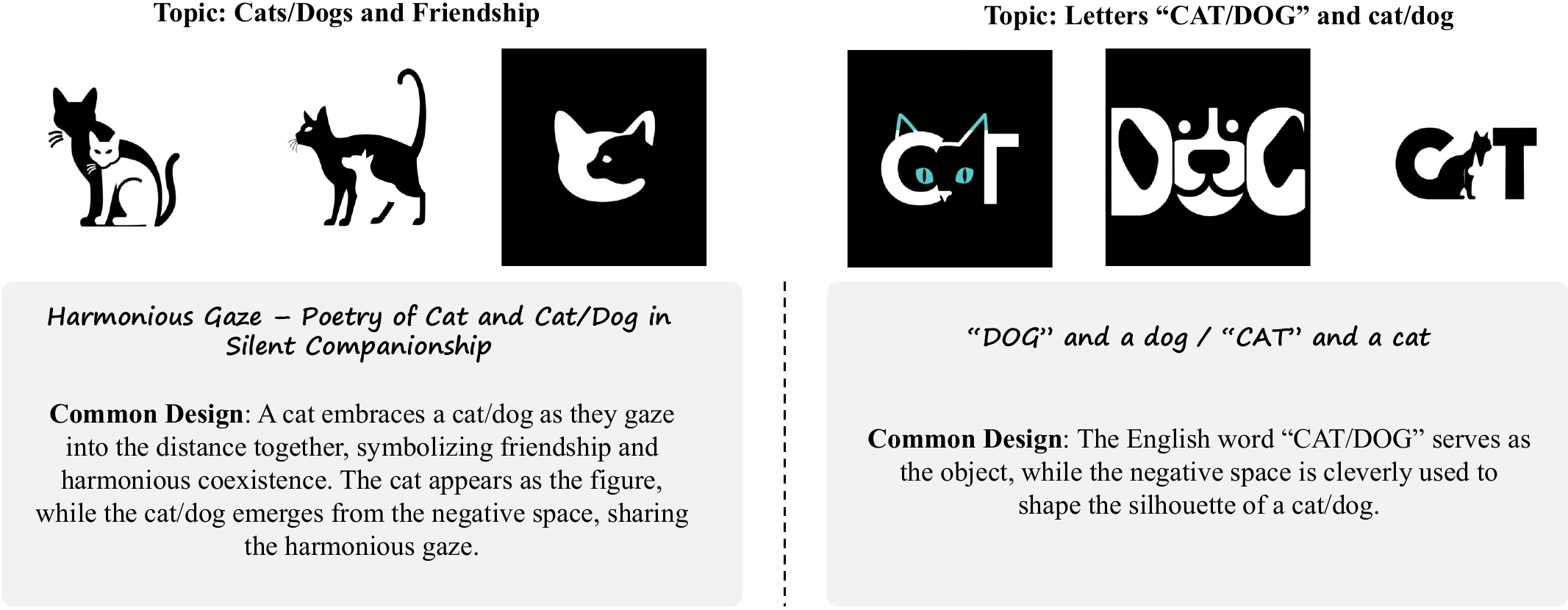}
    \label{fig:r2}
    
}
\caption{Positive-negative space composition results autonomously created by FaV-A. (Top) Examples demonstrating zero-shot semantic-to-spatial entanglement and structural coherence in multi-stable perception. (Bottom) Typography series showcasing cross-domain alignment and Gestalt emergence.}
\label{fig:results}
\vspace{-0.3cm}
\end{figure*}

\subsection{Discussions of Results by FaV-A}
Here we qualitatively evaluate the autonomous compositions generated by our FaV-A (see Figure \ref{fig:results}). Rather than merely viewing these outputs as static artistic artifacts, we regard them as both exquisite artistic works and as empirical evidence of the MLLM's capacity to execute complex zero-shot topological reasoning. In this discussion, we not only present the generated results but also provide in-depth analyses of the AI-generated design drafts that underpin these compositions. These results validate how the abstract duality of ``Form and Void'' is successfully grounded in precise visual geometries through our multi-modal reasoning pipeline, thereby imbuing the artworks with profound interpretive depth.
\vspace{-1em}
\paragraph{Zero-Shot Semantic-to-Spatial Entanglement}
\label{sec:results_void}
Traditional generative models typically struggle to map semantic intent to negative space, often defaulting to generating isolated foreground forms without explicit mask guidance. FaV-A, however, transforms spatial absence into a highly structured semantic entity. In \textit{Silent Duel: Between the Pieces and Their Masters} (see Figure~\ref{fig:r1} (1)), the negative space surrounding the chess piece is precisely sculpted into a human profile in the latent space. Similarly, in \textit{Phantom of Peace: A Doomsday Mushroom Cloud Beneath the Olive Branch} (see Figure~\ref{fig:r1} (3)), the mushroom cloud is synthesized entirely through the absence of surrounding textures rather than explicit contours. This demonstrates the agent's robust semantic-to-spatial entanglement, effectively translating the rich cognitive constraints of the semantic anchor ($\mathcal{P}_{\mathrm{anchor}}$) into complex topological boundaries without human-engineered layouts.
\vspace{-1em}
\paragraph{Structural Coherence in Multi-stable Perception}
\label{sec:results_ambiguity}
A fundamental technical challenge in generating visual illusions is maintaining structural coherence across overlapping visual elements without causing morphological collapse or semantic bleeding. FaV-A overcomes this by enforcing high-frequency edge preservation guided by its internal cross-modal attention mechanisms. In \textit{Whispered Kiss Beneath the Rose} (see Figure~\ref{fig:r1} (2)), a couple's interaction subtly emerges from the intrinsic curves of the rose, effectively blending surface geometry and hidden subtext. In \textit{Harmonious Gaze: Poetry of Cat and Cat/Dog in Silent Companionship} (see Figure~\ref{fig:r2} (Left)), both the primary silhouette and the embedded negative-space figures maintain high visual legibility. This confirms the agent's capacity to simultaneously optimize two distinct visual priors along a shared, continuous boundary, ensuring dual-semantics within a single structural contour.
\vspace{-1em}
\paragraph{Cross-Domain Alignment and Gestalt Emergence}
\label{sec:results_gestalt}
We further explore the agent's generalization capabilities in cross-domain scenarios involving typography. In the CAT/DOG typography series (see Figure~\ref{fig:r2} (Right)), text characters and animal silhouettes are fused into bistable configurations that resist immediate deterministic classification. For a standard vision model, textual forms and biological contours typically belong to disparate latent distributions, making seamless integration difficult. Yet, FaV-A successfully navigates this dual constraint: it interprets textual forms as the rigid ``ground'' while dynamically embedding animal silhouettes into the enclosed negative spaces. This algorithmic emergence of Gestalt perception highlights the MLLM's advanced compositional reasoning, proving its proficiency in handling conflicting multi-objective spatial tasks.

\section{Conclusion}
\label{sec:conclusion}
In this work, we introduced Form and Void Agent (FaV-A), a novel autonomous multi-modal system designed to synthesize complex positive-negative space compositions based on abstract textual topics. By formulating visual illusions as a sequence of high-level cognitive reasoning tasks, we bypass the heavy reliance on explicit spatial masks and structural templates prevalent in existing controlled diffusion models. Our proposed three-stage pipeline successfully integrates topic-driven semantic planning, in-context spatial anchoring, and mask-free \texttt{(I\&T) to (I\&T)} synthesis. Experimental results demonstrate that FaV-A not only executes cross-modal visual alignment with high structural coherence but also exhibits advanced zero-shot topological reasoning capabilities, such as multi-stable perception and cross-domain typographic integration. Ultimately, FaV-A highlights the immense potential of Large Multimodal Models to serve as autonomous cognitive partners in high-level aesthetic design, paving the way for more sophisticated, semantic-driven computational artistry.

\section*{Acknowledgements}
This work was supported in part by the National Natural Science Foundation of China under no. 62572458, and in part by the Shanghai Pujiang Talent Program under no. 25PJC081.

{
    \small
    \bibliographystyle{ieeenat_fullname}
    \bibliography{main}
}


\end{document}